# Modelling Visual Semantics via Image Captioning to extract Enhanced Multi-Level Cross-Modal Semantic Incongruity Representation with Attention for Multimodal Sarcasm Detection

Sajal Aggarwal, Ananya Pandey, Dinesh Kumar Vishwakarma*

Biometric Research Laboratory, Department of Information Technology, Delhi Technological University, Bawana Road, Delhi-110042, India

sajalaggarwal_it20b11_08@dtu.ac.in , ananyapandey_2k21phdit08@dtu.ac.in, dinesh@dtu.ac.in*

**Abstract**

Sarcasm is a type of irony, characterized by an inherent mismatch between the literal interpretation and the intended connotation. Though sarcasm detection in text has been extensively studied, there are situations in which textual input alone might be insufficient to perceive sarcasm. The inclusion of additional contextual cues, such as images, is essential to recognize sarcasm in social media data effectively. This study presents a novel framework for multimodal sarcasm detection that can process input triplets. Two components of these triplets comprise the input text and its associated image, as provided in the datasets. Additionally, a supplementary modality is introduced in the form of descriptive image captions. The motivation behind incorporating this visual semantic representation is to more accurately capture the discrepancies between the textual and visual content, which are fundamental to the sarcasm detection task. The primary contributions of this study are: (1) a robust textual feature extraction branch that utilizes a cross-lingual language model; (2) a visual feature extraction branch that incorporates a self-regulated residual ConvNet integrated with a lightweight spatially aware attention module; (3) an additional modality in the form of image captions generated using an encoder-decoder architecture capable of reading text embedded in images; (4) distinct attention modules to effectively identify the incongruities between the text and two levels of image representations; (5) multi-level cross-domain semantic incongruity representation achieved through feature fusion. Compared with cutting-edge baselines, the proposed model achieves the best accuracy of 92.89% and 64.48%, respectively, on the Twitter multimodal sarcasm and MultiBully datasets.

## 1 Introduction

Sarcasm is a kind of wit used to poke fun at or make disparaging remarks about a person or situation. The literal interpretation of sarcastic remarks frequently suggests the polar opposite of their intended meaning. This makes sarcasm detection a somewhat onerous task. For instance, "Oh, I love it when people chew loudly. It's such a pleasant sound." is a remark wherein words indicative of a positive sentiment, namely 'love' and 'pleasant', are used to express an unfavourable opinion for the 'unpleasant' behaviour of 'chewing loudly'. Sarcasm can be communicated by an amalgamation of verbal and nonverbal cues, including but not limited to alterations in vocal tone, exaggeration of a particular word, elongation of a syllable, or a neutral facial expression while delivering a seemingly positive remark. Traditional methods for detecting sarcasm have primarily relied on analyzing either textual [1], [2] or audio inputs [3] alone. However, the fact that sarcasm is fundamentally characterized by linguistic and contextual incompatibility has necessitated the shift to multimodal datasets [4]–[7] so as to capture meaningful information from multiple modalities and contextual history.

The sarcasm in the tweet "I love commuting", for example, is lost if the accompanying photograph of a congested traffic situation is disregarded. The satirical tone of the tweet can only be fully appreciated when both text and image are taken into account. Ignoring the visual cues could lead one to mistakenly interpret the comment as sincere, missing the satirical undertones entirely. **Figure 1** provides additional clarification on the necessity of incorporating multimodal inputs in order to achieve accurate sarcasm detection. This is demonstrated through the inclusion of instances from two benchmark datasets. Hence, to improve the complexity and accuracy of an automated system for detecting sarcasm, it is crucial to go beyond the limitations of textual data and incorporate additional data from different modalities that capture a wide range of information, enabling the identification of sarcasm.

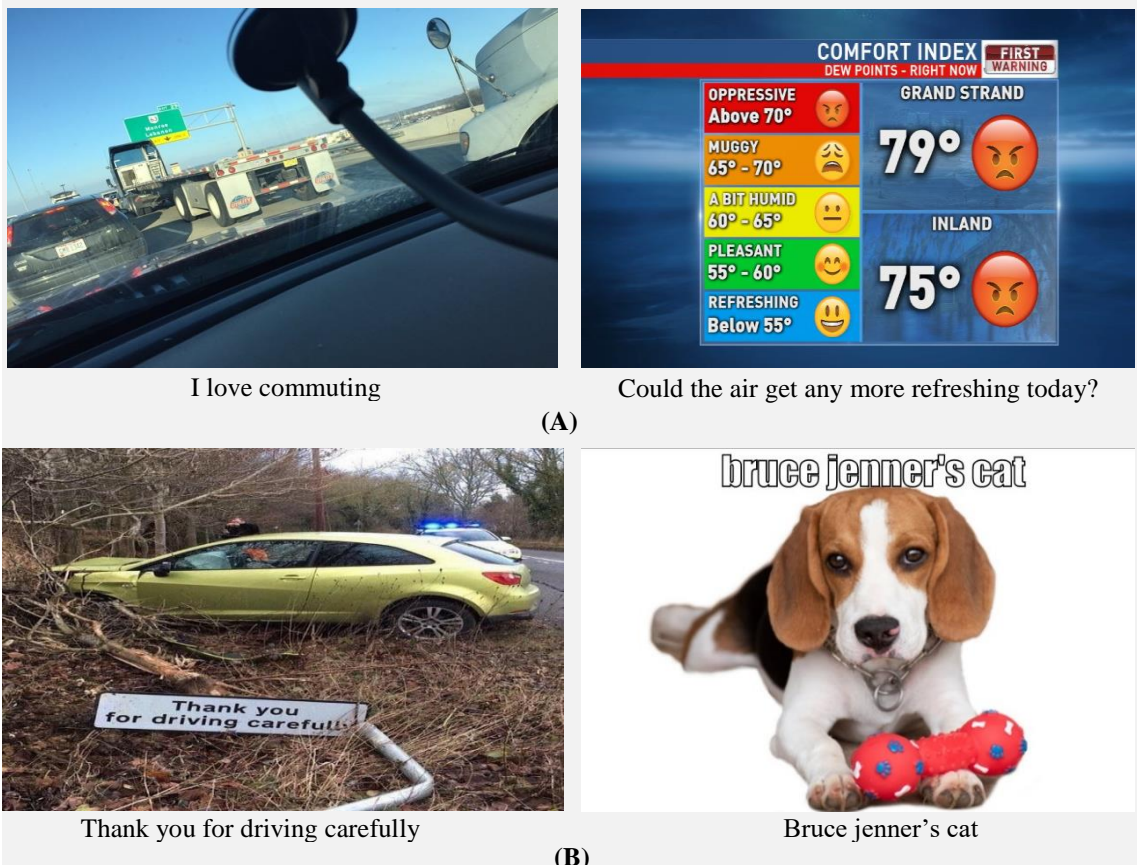

**Figure 1** Examples from the Twitter multimodal sarcasm (A) and MultiBully (B) dataset demonstrate the limitations of relying solely on textual data to accurately convey sarcastic intent in certain instances.

Studies have been conducted to extract image attributes [6] or adjective-noun pairs (ANPs) [8] from images for multimodal sarcasm detection in image-text pairs. The objective of these studies is to incorporate an additional modality to establish a correlation between textual and visual information, thereby enhancing the understanding of visual semantics. However, the incorporation of such elements in the analysis can, at times, lead the sarcasm detection model astray. This occurs when these elements generate attributes or objects that fail to capture the fundamental essence of the visual semantics. In other words, they may refer to objects or attributes that do not effectively support the sarcasm detection task or highlight the inherent incongruity between the visual and the text. As a result, this constraint has the capacity to impede the efficacy of sarcasm detection. In order to upgrade the accuracy and coherence of image descriptions and to bring out the disconnect between textual and visual content, we introduce a novel framework that utilises a pre-trained encoder-decoder framework to generate detailed image captions. These captions serve as an additional modality for detecting sarcasm in multimodal contexts. **Figure 2** serves to enhance the clarity of differentiating between the utilisation of image attributes and image captions. It accomplishes this by presenting examples that demonstrate how visual attributes, in contrast to captions, fail to adequately capture significant information pertaining to the image, thereby hindering the detection of sarcasm.

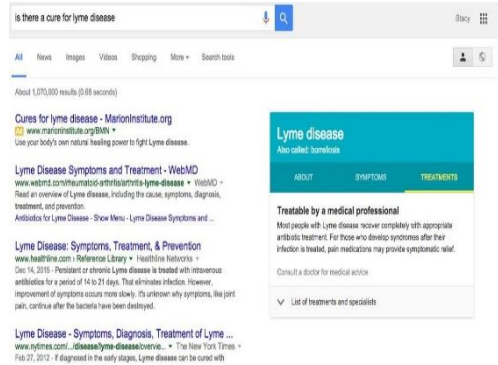

**Text:** Please tell me more about the magic cure you found for # lymedisease

**Attributes:** white, scene, sign, book, screen

**Caption:** A google search page shows a website with Lyme disease

**Sarcasm**

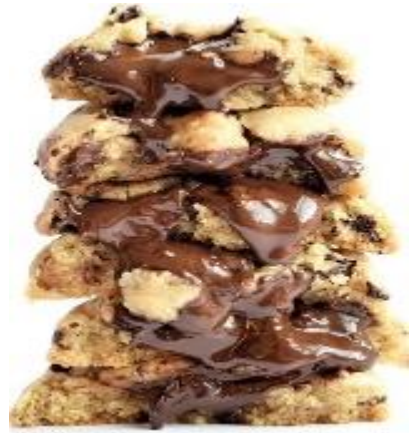

**Text:** Chocolate chip cookies

**Attributes:** dark, brown, nuts, black, food

**Caption:** A stack of chocolate chip cookies with a chocolate drizzle

**Non-Sarcasm**

**Figure 2** In contrast to attributes, image captions in the above examples can improve visual semantics identification for sarcasm detection. The image captioning module used in this study has strong reading comprehension, allowing it to produce accurate descriptions even for images with text.

**Major Contributions**

The major contributions of the present study are laid out as follows:

1. A cross-lingual language model is employed as the textual tokenizer and encoder to acquire the feature representations for both the text data and the image captions. The efficacy of this model is notable as it has undergone extensive pretraining on a wide array of data encompassing more than 100 languages, including Hindi. This is particularly advantageous as the MultiBully dataset consists of code-mixed English-Hindi statements.
2. In contrast to the majority of prior research studies [6], [9]–[14], which typically utilize a variant of the ResNet architecture for generating image feature representations, the proposed framework employs a more robust self-regulated residual ConvNet [15] architecture to encode images into feature vectors. In order to enhance the obtained feature maps and prioritize the most important regions in the image, we incorporate a lightweight spatially aware attention module into our image feature extraction branch.

3. As opposed to other studies [6], [8], [14], which utilize either image attributes or objects as an additional modality, our proposed approach offers a more efficient method by integrating descriptive captions generated through a generative image-to-text transformer model. This integration allows for the incorporation of an additional source that emphasizes the discrepancy between modalities, ultimately enhancing the accuracy of sarcasm detection.
4. To encapsulate the contextual incongruity between the input text and the high-level image description generated through captions, a co-attention layer is integrated on top of the transformer-based textual feature extraction and caption feature extraction branches. The inter-modality discrepancy between the textual and low-level image representations is effectively addressed through the utilization of a multi-head self-attention layer. Therefore, we are able to capture the incongruity between the textual and visual data in two distinct manners, operating at separate levels. To the best of our knowledge, we are the first to propose a resilient and improved multi-level representation of inter-modal semantic incongruity based on descriptive image captions for multimodal sarcasm detection.
5. Extensive empirical investigations on two publicly accessible datasets [6], [16] demonstrate that the proposed framework exhibits a consistent enhancement when compared to baseline approaches and achieves a competitive level of performance in comparison to several established multimodal sarcasm detection techniques.

**Organization:** The ensuing sections of this document are organised in the following format: **Section 2** classifies the prior research carried out in the domain of sarcasm detection according to the various data modalities employed. In this section, various subsections elaborate on the cutting-edge techniques utilized for sarcasm recognition in both unimodal and multimodal contexts. **Section 3** provides an in-depth explanation of the proposed framework for detecting sarcasm using image-text pairs. **Section 4** provides an overview of the dataset characteristics for the two datasets employed in this research. **Section 5** offers a detailed description of the experimental setups and presents the findings obtained from the application of the proposed framework. **Section 6** provides a summary of the current study and offers suggestions for potential areas of investigation in future research.

## 2 Literature Review
### 2.1 Unimodal Sarcasm Detection
In this section, several state-of-the-art sarcasm detection approaches based on a single modality have been discussed.

#### 2.1.1 Textual Data
Rule-based methodologies endeavor to recognize sarcasm in textual data by means of discerning particular pieces of evidence. According to Maynard and Greenwood [17], the sentiment expressed through hashtags serves as a pivotal determinant for detecting sarcasm. Hashtags are frequently employed by authors of tweets to accentuate sarcasm. Consequently, if the emotion conveyed by a hashtag failed to align with the overall content of the tweet, it was suggested that the tweet was intended to convey sarcasm. For statistical sarcasm detection, the majority of methodologies utilized bag-of-words as their primary feature representation. As demonstrated by Joshi et al. in [18], a crucial component of sarcasm detection is the use of word embeddings as a basis for similarity evaluation. Sarcasm detection studies based on conventional techniques extracted a myriad of features, including emoticons, unigrams [2], n-grams [19], POS sequences [20], semantic similarity [21], sentiment [22], and word frequency [23].

With the increasing prevalence of deep learning-based architectures in the realm of NLP, researchers have employed these methodologies to tackle the intricate challenge of automatic sarcasm detection [18], [24]–[29]. These studies delved into the intricate nuances of contextual elements, such as the complex interaction between the content of tweets and user reactions. In [25], Amir et al. introduced a novel architecture that utilized CNNs to acquire user embeddings alongside utterance-based embeddings to acquire user-specific contextual information. In [26], Poria et al. examined the use of deep convolutional networks as a means of discerning sarcasm in tweets. Goel et al. [30] trained an ensemble model comprising CNN, LSTM and GRU to effectively identify sarcastic statements in two benchmark datasets. Misra and Arora [29] proposed an attention-based hybrid neural network architecture to detect sarcasm in news headlines. Given that an ample supply of annotated data necessitates a substantial expenditure of human effort, Wang et al. [31] examined sarcasm detection from an unsupervised standpoint. The authors proposed a masking and generation paradigm to extract inconsistencies and acquire knowledge pertaining to the expression of sarcasm.

### 2.1.2 Acoustic Data

Tepperman et al. [3] conducted an experiment to examine the recognition of sarcasm in spoken dialogue. They analyzed 131 instances of the phrase "yeah right", which were obtained from the Switchboard and Fisher corpora [32], [33]. The authors investigated multiple cues, encompassing prosodic, spectral, and contextual factors, and provided evidence that the inclusion of prosodic features is dispensable in the detection of sarcasm as long as spectral and contextual cues are considered.

### 2.2 Multimodal Sarcasm Detection

Multimodal sarcasm detection seeks to identify the sarcastic expression across several modalities, particularly image-text pairs [4] or videos [7], as opposed to relying solely on text-based or audio-based sarcasm recognition. The following subsections provide an elaboration of various approaches utilized for the detection of sarcasm in multimodal contexts.

### 2.2.1 Text and Visual Data

Sarcasm detection systems that depend solely on textual analysis may struggle to distinguish between genuine and sarcastic speech. Given the prevalence of text-image amalgamations within contemporary social media platforms, multimodal approaches that capture the disparity between the two modalities seem to be more promising for sarcasm prediction. Schifanella et al. [4] were the first to approach sarcasm identification as a multimodal classification problem by compiling a dataset comprising text-image posts from three social media platforms - Twitter, Tumblr and Instagram. They performed sarcasm recognition by concatenating image and text features that were either manually handcrafted or deep-learning-based. Using a combination of text, numeric, and visual information taken from Facebook posts, Das and Clark [34] trained multiple ML classifiers to identify sarcasm. Sangwan et al. [5] created two multimodal sarcasm recognition datasets: the Silver-Standard dataset, consisting of 10K sarcastic and non-sarcastic Instagram posts each, categorized on the basis of hashtags, and the Gold-Standard dataset, which includes 1600 randomly selected sarcastic posts from the first dataset and annotated first using only the text modality, and then reannotated using both modalities. The authors developed an RNN-based deep learning system to identify sarcasm by capitalizing on the interdependence of text and visuals.

An extensive dataset for multimodal Twitter sarcasm detection was introduced by Cai et al. in [6]. The authors put forward a hierarchical fusion framework that combined text, image, and attribute feature representations extracted using BiLSTM, ResNet-50, and ResNet-101 models, respectively. Pandey and Vishwakarma [35], [36] employed attention mechanisms in conjunction with convolutional neural networks to carry out sentiment analysis on multimodal datasets comprising image-text pairs. In [10], Wang et al. utilized the pre-trained BERT and ResNet models for extracting the text and image feature representations, respectively and established a connection between the vector spaces of BERT and ResNet using a bridge layer. In order to represent the differences between the text and the images, the authors used multi-head attention and devised a 2D-intra-attention mechanism to draw focus to the disparities. Xu et al. [8] proposed a novel D&R Net (Decomposition and Relation Network) to model cross-modality incongruity and semantic association for sarcasm detection. While the decomposition network focused on contextual contrast between the text and image in high-level spaces, the relation network used a variant of multi-head attention to capture the degree of semantic association between the input text and adjective-noun pairs extracted from the image.

Zhao et al. [11] utilized BERT and Bi-GRU to extract text features and pre-trained ResNet-152 to extract image features. They then employed textual and visual attention mechanisms to derive context vectors, prioritizing specific keywords and image patches for enhanced focus. The coupled-attention model, as proposed, proceeded to calculate fusion memory vectors for k iterations. In this process, a memory vector, indexed by iteration i, stored the combined visual and textual information that had been gathered up to that particular iteration. Ultimately, the final memory vector underwent processing via a single-layer softmax classifier in order to detect sarcasm.

In [37], Liang et al. explored the use of graph convolution networks to jointly learn the sentiment incongruity within and across the text and image modalities as captured by in-modal and cross-modal graphs, respectively. In [38], Liang et al. presented a novel cross-modal graph convolutional network (CMGCN) to identify key parts of an image and then correlate them with the textual tokens to detect sarcasm.

Liu et al. [39] presented an innovative neural model that incorporated logic-based techniques to detect instances of multimodal rumors and sarcasm. The model utilized interpretable logic clauses to effectively articulate the reasoning process involved in the target task. A new framework called MuLOT (Multimodal Learning using Optimal Transport) was proposed by Pramanick et al. in [12] to detect sarcasm and humor. This framework uses self-attention to consider associations within a single modality, optimal transport to utilize cross-modal relationships, and multimodal attention fusion to account for inter-dependencies between modalities.

A multimodal meme dataset - "MultiBully", annotated with labels for bullying, sentiment, sarcasm, and emotion, was developed and provided by Maity et al. in [16]. The authors concluded that the proposed multimodal multitask frameworks for recognition of cyberbullying and the three supplemental classification problems (sarcasm detection, sentiment analysis, and emotion recognition) prevailed over their unimodal and single-task variants, thus boosting the accuracy of cyberbullying detection.

### 2.2.2 Text and Acoustic Data

Bedi et al. [40] proposed a new Hindi-English code-mixed dataset, MaSaC, for multimodal sarcasm detection in conversational dialogue comprising 15K utterances from 400 scenes of the Indian TV comedy show 'Sarabhai v/s Sarabhai'. The authors also developed MSH-COMICS, an attention-based multimodal classification model consisting of a hierarchical utterance-level attention module in order to acquire a more detailed linguistic representation of each statement.

### 2.2.3 Text, Visual and Acoustic Data

The first video dataset for sarcasm detection, MUStARD, was released by Castro et al. in [7]. The authors collected and annotated 690 audio-visual utterances from popular TV shows and used an SVM classifier to predict sarcasm using a combination of textual, audio and visual features. Chauhan et al. [41] used emotion and sentiment for sarcasm recognition after updating the MUStARD dataset to incorporate relevant emotion and sentiment labels. By adding fresh utterances from similar sources, Ray et al. [42] doubled the size of the extended MUStARD dataset provided by Chauhan et al. Labels for arousal, valence, and sarcasm type were added to each phrase, and some incorrectly assigned emotions were also fixed. This revised dataset was released by the authors under the name MUStARD++. The dataset was benchmarked using different combinations of pre-trained feature extraction algorithms and multimodal fusion techniques. Chauhan et al. [43] introduced an enhanced version of the MUStARD dataset, known as SEEmoji MUStARD, wherein each utterance was annotated with a pertinent emoji, along with the associated emotion and sentiment conveyed by the emoji.

### 2.3 Research Limitations and Motivation

Despite the extensive study of sarcasm recognition in textual data, multimodal sarcasm detection is still a developing discipline. To evaluate their approach, most studies based on multimodal sarcasm recognition in image-text pairs use the Twitter multimodal sarcasm detection dataset. In order to stimulate more research on similar datasets, this paper demonstrates its findings on both the Twitter dataset and another less-studied dataset, MultiBully. The incorporation of this supplementary dataset in our research prompted us to utilize a multilingual variant of a transformer module to handle code-mixed statements. This is in contrast to previous studies that predominantly utilized the standard BERT model to extract features for English-language statements in the Twitter dataset. The majority of studies discussed in **section 2.2.1** have employed one or the other variant of the ResNet architecture as the conventional approach for visual feature extraction. However, there has been a lack of exploration into alternative methods for enhancing the robustness of visual feature extraction. Inspired by this observed pattern, we propose a more robust visual feature extraction pipeline. This pipeline incorporates a self-regulated ConvNet combined with a spatially aware attention module. This integration imposes minimal computational burden while simultaneously improving the quality of the extracted feature maps. Image attributes have been utilized in multimodal sarcasm detection research to incorporate an extra modality for improving the detection of incongruity. The inclusion of supplementary external knowledge is crucial for emphasizing the disparity between the textual message and the visual data. However, it is imperative that this knowledge source is more comprehensive and provides clearer explanations. In certain instances, attributes of images can provide meaningful information, particularly when the images depict simple scenes and standard objects. However, these attributes are insufficient in cases where the image holds a deeper meaning, particularly when there is text embedded within the image that must be understood in order to accurately interpret the semantic information conveyed by the image. Motivated by this shortcoming, this study suggests utilizing a more reliable external source of knowledge, which involves using descriptive image captions generated by an advanced encoder-decoder architecture. This architecture is capable of understanding text embedded in images and producing meaningful captions that accurately represent the visual semantics. This presents an additional opportunity for us to capture the differentiation between the given text and image, thereby improving the precision of automated multimodal sarcasm detection.

### 3 Proposed Methodology

In this section, we present an extensive account of the proposed framework for multimodal sarcasm detection in image-text pairs. **Figure 3** depicts the basic building blocks of the proposed architecture, which include the following: 1) A transformer-based textual feature extraction branch to capture the latent representation of the text modality; 2) An attentional visual feature extraction branch comprising [15] combined with a lightweight attention module to uncover the image modality's latent representation; 3) An encoder-decoder architecture to generate descriptive caption corresponding to each image, followed by conversion to feature representation via a cross-lingual language model; and 4) Two distinct intermodal attention modules in order to effectively capture and model the inconsistencies that exist between textual information and two different levels of image representation. The pseudocode for the proposed approach has been outlined in **Table 1**.

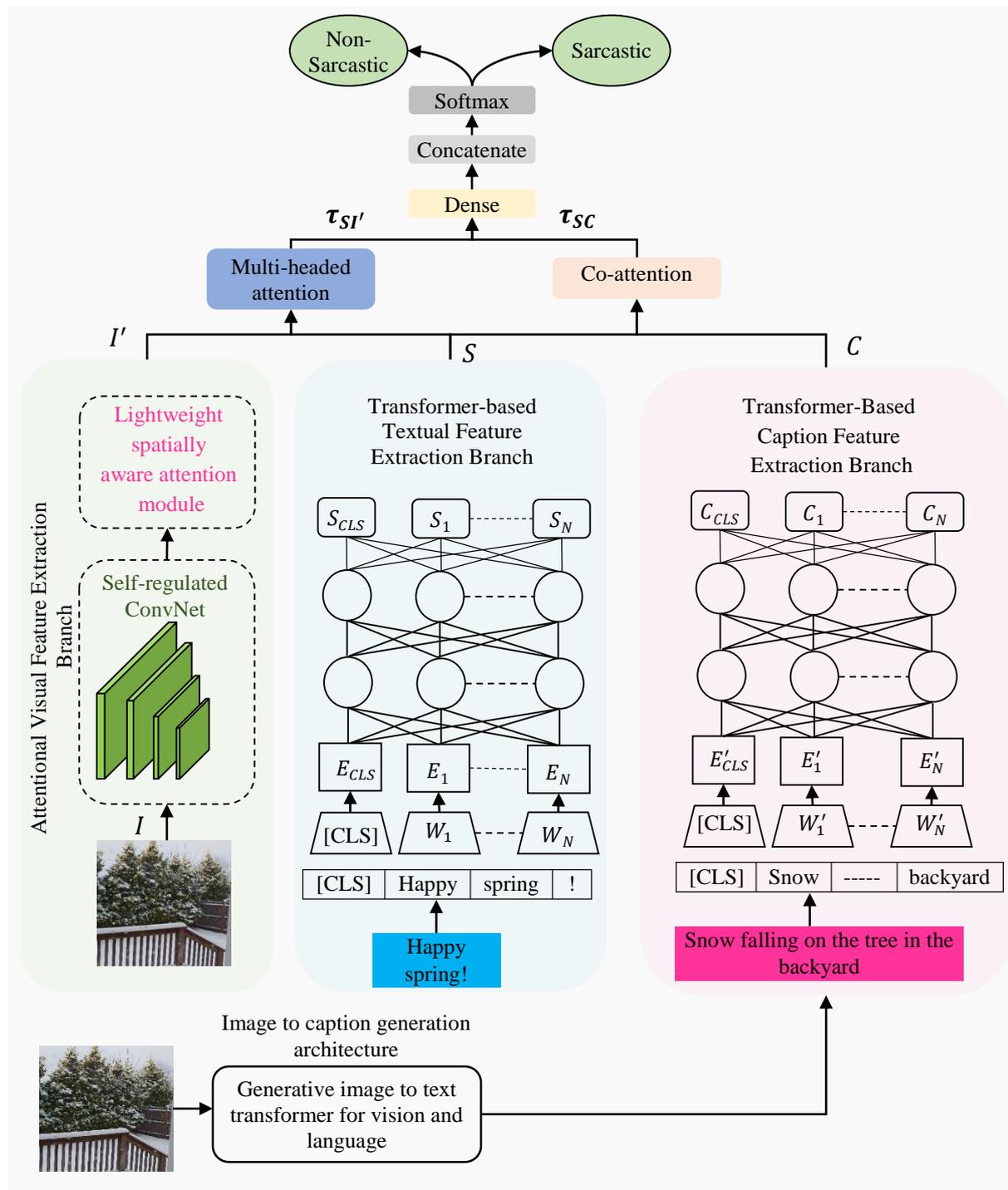

**Figure 3** Proposed framework

**Table 1** Algorithm for the proposed model

| |
|---|
| **Algorithm-1:** Enhanced Multi-Level Cross-Domain Incongruity Representation using Image Captions for Multimodal Sarcasm Detection |
| **Input:** Multimodal inputs, $M = \{W, I, W'\}$, where $W, I$, and $W'$ denote text, image, and caption correspondingly |
| **Output:** Sarcasm detection label, $Y \in \{non-sarcastic, sarcastic\}$ |
| for $\mathbb{E} \leftarrow 1$ to epochs do<br>    $W_{tok} = \{[CLS], w_1, w_2, w_3, \dots, w_t\} \leftarrow W = \{w_1, w_2, w_3, \dots, w_t\}$ word sequence tokenization using **Eq. (2)**<br>    $S \leftarrow W_{tok} = \{[CLS], w_1, w_2, w_3, \dots, w_t\}$ text feature representation using **Eq. (3)**<br>    $ConvNet_{self-reg}(I) \leftarrow I$ image feature representation using **Eq. (4)**<br>    $ConvNet_{self-reg}(I)_{transformed} \leftarrow ConvNet_{self-reg}(I)$ projection of image features into same dimensions as text features using **Eq. (5)**<br>    $I' \leftarrow ConvNet_{self-reg}(I)_{transformed}$ spatially selective feature maps using **Eq. (6)**<br>    $W' \leftarrow I$ using generative image-to-text transformer model<br>    $W'_{tok} = \{[CLS], w'_1, w'_2, w'_3, \dots, w'_u\} \leftarrow W' = \{w'_1, w'_2, w'_3, \dots, w'_u\}$ caption tokenization using **Eq. (7)**<br>    $C \leftarrow W'_{tok} = \{[CLS], w'_1, w'_2, w'_3, \dots, w'_u\}$ caption feature representation using **Eq. (8)**<br>    $\tau_{SI'} \leftarrow (S, I')$ text-image incongruity representation using **Eq. (9), Eq. (10), Eq. (11),** and **Eq. (12)**<br>    $\tau_{SC} \leftarrow (S, C)$ text-caption incongruity representation using **Eq. (13), Eq. (14), Eq. (15),** and **Eq. (16)**<br>    $F \leftarrow (\tau_{SI'}, \tau_{SC})$ fused incongruity representation using **Eq. (17)**<br>    $y \in Y \leftarrow F$ model prediction using **Eq. (18)**<br>end for |

## 3.1 Task Definition

In the context of this study, multimodal sarcasm detection seeks to ascertain whether a particular text-image pair is intended to convey sarcasm or not. In the selected dataset $D$, each sample $d \in D$ comprises an input text sequence of words, denoted as $W = \{w_1, w_2, w_3, \dots, w_t\}$, along with a corresponding image $I$. To enhance the comprehension of image semantics, we employ a more effective approach by generating a descriptive caption corresponding to each image. Therefore, our model is capable of accepting additional input as an image caption, $W'$. Thus, the input triplets accepted by the model can be represented as shown in ***Equation*** (**1**).

$$Multimodal\ Input, M = \{Text\ (W), Image\ (I), Image\ Caption\ (W')\} \qquad (1)$$

Our objective is to assign a label $y \in Y$ to each input triplet $M$, where $Y = \{non-sarcastic, sarcastic\}$.

## 3.2 Transformer-based Textual Feature Extraction Branch

The cross-lingual language model [44] has been employed to get the textual features from input text sequence of words, $W = \{w_1, w_2, w_3, \dots, w_t\}$. The model tokenizer, $\mathbb{X}_{tokenizer}$ is utilized as an embedding module to generate text embeddings that encompass extensive semantic information. To achieve this, the input sequence is initially tokenized into a specific format, as shown in ***Equation*** (**2**).

$$W_{tok} = \mathbb{X}_{tokenizer}(W) = \{[CLS], w_1, w_2, w_3, \dots, w_t\} \qquad (2)$$

where, $w_i \in \mathbb{R}^d$ and $d$ denotes the embedding size. The sequential output, $S$ obtained from the use of model encoder, $\mathbb{X}_{encoder}$ [44] for all tokens in the input $W_{tok}$ can be denoted as shown in ***Equation*** (**3**).

$$S = \mathbb{X}_{encoder}(W_{tok}) \tag{3}$$

where, $S \in \mathbb{R}^{T \times d}$, where $d = 768$ stands for the dimension of each token, and $T$ represents the maximum total sequence length.

With over 2 terabytes of cleaned CommonCrawl data [45], the masked language model [44] has already been pre-trained on over a hundred languages, including Hindi. We chose this particular variant of [46] for our investigation since the majority of the input sentences in the 'MultiBully' meme dataset are code-mixed Hindi-English. By fixing its weaknesses with a technique called sentence piece tokenization, [44] notably outperforms other multilingual models, such as [47], on several cross-lingual benchmarks.

### 3.3 Attentional Visual Feature Extraction Branch
The ensuing subsections comprehensively elucidate the proposed methodology for capturing vital information from the visual content.

#### 3.3.1 Self-regulated Residual ConvNet
Upon receiving an input image, $I$ with dimensions $I_H \times I_W$, the initial step involves resizing the image to dimensions 224×224. Subsequently, the image is partitioned into regions of size 7×7. Following that, each region is passed through the [15] model, as illustrated in **Figure 4**, to obtain a regional feature representation. After eliminating the terminal FC layer, the output of final convolutional layer corresponding to input image I can be denoted as depicted in ***Equation* (4).**

$$ConvNet_{self-reg}(I) \in \mathbb{R}^{r \times 2048} = \{r_i\}_{i=1}^{49} \tag{4}$$

where $r_i \in \mathbb{R}^{2048}$ represents the 2048-dimensional feature representation corresponding to a single image region.

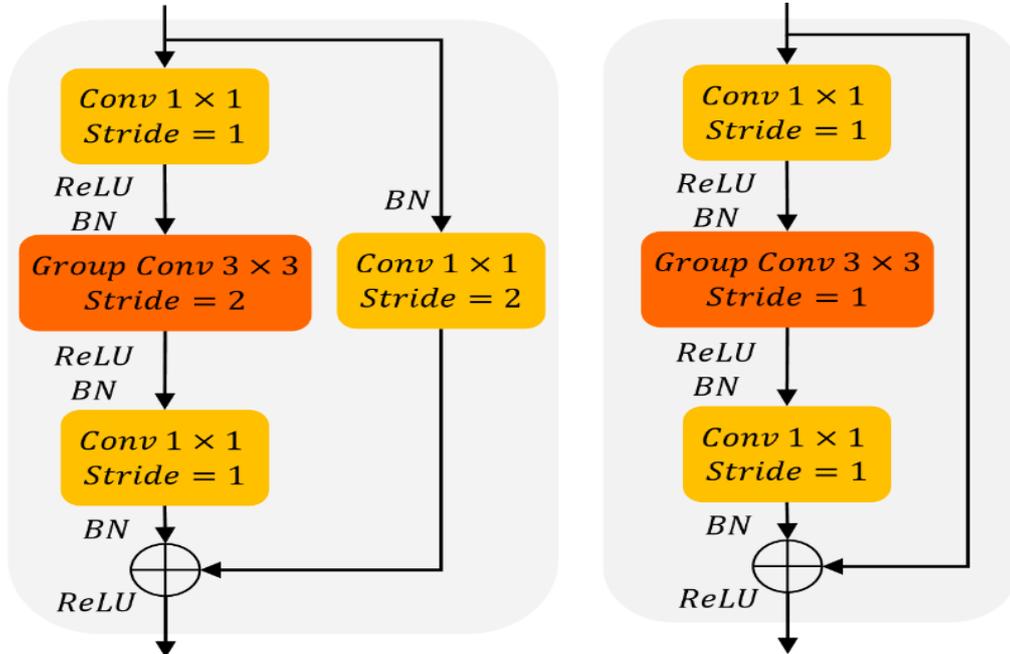

**Figure 4** Structures of self-regulated residual ConvNet blocks

The architecture in [15] introduced a regulator module consisting of convolutional RNNs to extract complementary features and control the necessary memory information being passed to the next block. The input feature from the $jth$ building block, $Ij$, together with the sequential information, is encoded by a recurrent unit of the RNN (LSTM or GRU) to produce the hidden state $H_j$. To produce the output $O_j$, the hidden state is concatenated with $I_j$ and fed to the next convolution operation and recurrent unit.

To ensure that the extracted image features are projected into the same dimensions as those of the text feature representation, the encoded image representation is subjected to a linear transformation as shown in **Equation** (5).

$$ConvNet_{self-reg}(I)_{transformed} \in \mathbb{R}^{r \times d} = ConvNet_{self-reg}(I)L_V \qquad (5)$$

where, $L_V \in \mathbb{R}^{2048 \times d}$ signifies a trainable parameter, and d corresponds to the dimension of each token of the textual sequence.

### 3.3.2 Lightweight Spatially Aware Attention Module

Attention modules are extensively employed in tasks related to computer vision. The primary intent of this integration is to assist models in discerning particular areas that necessitate specific focus. Nevertheless, the incorporation of attention mechanisms is somewhat constrained as lightweight networks are unable to effectively manage the associated computational load. Hence, to enhance the accuracy of image feature extraction, this study integrates a simple and flexible attention mechanism [48] that imposes a minimal computational load. The conventional approach of channel attention enables convolutional networks to acquire about the specific aspects that require emphasis during the learning process. As opposed to channel attention, which uses two-dimensional global pooling to reduce a feature tensor to a feature vector and disregards the importance of incorporating positional information, [48] splits global pooling operation into separate one-dimensional feature encoding transformations in the horizontal and vertical directions, so as to efficiently incorporate spatial coordinate data into the resulting attention maps. As a result, precise positioning information can be maintained along one spatial dimension while long-range dependencies can be captured along the other. A pair of attention maps that take into account both direction and location are then encoded from the resulting feature maps to further refine representations of the input feature map's notable features. The integration of the attention module, depicted in **Figure 5**, with blocks in convolutional networks allows for the enhancement of features by zeroing in on the most crucial aspects of the input feature representations, owing to its adaptability and low overhead.

Let the attention module's input feature tensor be $\mathbb{X} \in \mathbb{R}^{\mathbb{C} \times \mathbb{H} \times \mathbb{W}}$, where $\mathbb{C}$ denotes number of channels in $\mathbb{X}$, $\mathbb{H}$ symbolizes its height, and $\mathbb{W}$ stands for its width. Then the output $I' \in \mathbb{R}^{\mathbb{C} \times \mathbb{H} \times \mathbb{W}}$, represents a transformed tensor of the same size as $\mathbb{X}$ but with enhanced representations. To precisely capture inter-channel associations and positional information, the applied attention mechanism operates through a two-step process. This process involves information embedding, where the relevant data is encoded, and attention generation, where the attention weights are computed. The first step performs two transformations, one along each of the two spatial directions, to model the long-range interactions spatially while preserving the positional information. Each channel is encoded along the height and width by applying two average pooling kernels, ($\mathbb{H}$,1) and (1,$\mathbb{W}$). The second phase aims at effectively capturing inter-channel relationships from the aggregated feature maps generated in the preceding stage.

The two feature maps are initially combined and subsequently put through a shared 1×1 convolutional transformation, thereby yielding an intermediate feature map, $\mathbb{f} \in \mathbb{R}^{\frac{\mathbb{C}}{\mathbb{r}} \times (\mathbb{H}+\mathbb{W})}$, that captures spatial information along both horizontal and vertical directions. In this context, 𝕣 denotes the reduction ratio, which serves as the controlling factor for determining the block size. This is followed by factorizing $\mathbb{f}$ into two separate feature maps $\mathbb{f}^{\mathbb{h}} \in \mathbb{R}^{\frac{\mathbb{C}}{\mathbb{r}} \times \mathbb{H}}$, and $\mathbb{f}^{\mathbb{w}} \in \mathbb{R}^{\frac{\mathbb{C}}{\mathbb{r}} \times \mathbb{W}}$, which are then transformed into corresponding tensors $\mathbb{g}^{\mathbb{h}}$ and $\mathbb{g}^{\mathbb{w}}$ with the same number of channels as $\mathbb{X}$ via $1 \times 1$ convolutional transformations and sigmoid functions. As depicted in ***Equation*** (**6**), the output tensor $I'$ of the attention module is calculated using $\mathbb{g}^{\mathbb{h}}$ and $\mathbb{g}^{\mathbb{w}}$ as attention weights.

$$I' = \mathbb{X} \times \mathbb{g}^{\mathbb{h}} \times \mathbb{g}^{\mathbb{w}} \tag{6}$$

where, $\mathbb{X}$ stands for the image feature representation, $ConvNet_{self-reg}(I)_{transformed}$, as derived in ***Equation*** (**5**). The attention maps, $\mathbb{g}^{\mathbb{h}}$ and $\mathbb{g}^{\mathbb{w}}$, which possess the ability to perceive direction and sensitivity to position, work in tandem on the input feature map to strengthen the information pertaining to the focal points.

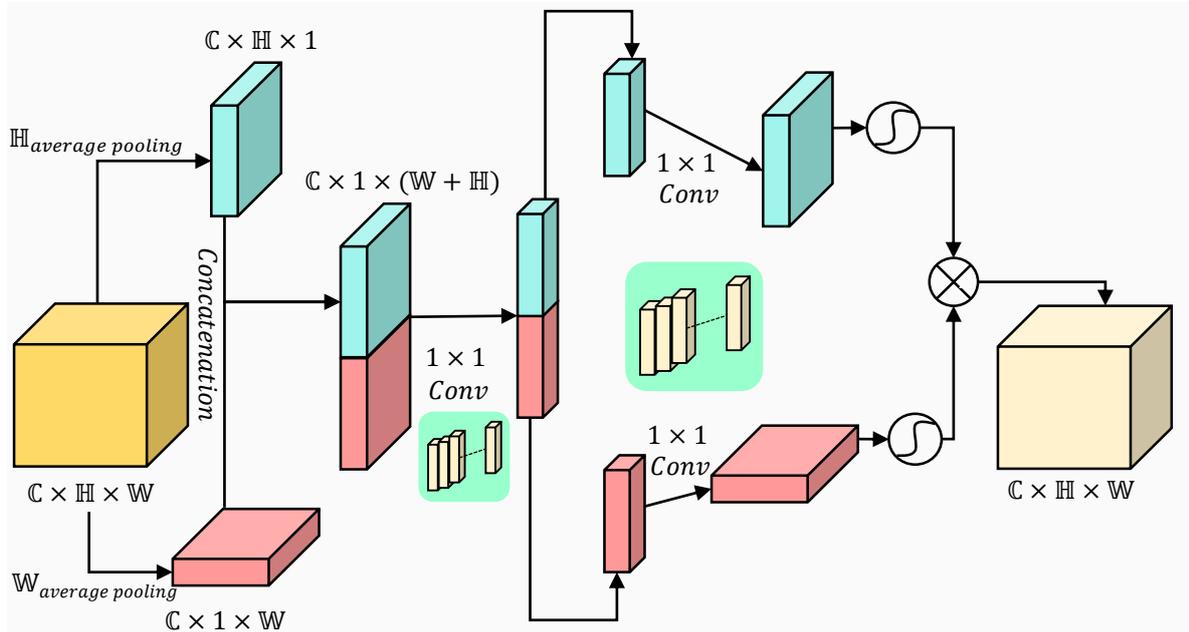

**Figure 5** Structure of the lightweight attention module integrated to obtain spatially selective feature maps

### 3.4 Image Captioning
The following subsections shed light on the architecture utilized for image captioning and the process of feature extraction for the generated captions.

#### 3.4.1 Encoder-Decoder Architecture
The concept of image captioning pertains to the procedure of generating descriptive captions, or textual representations, derived from input images. The generation of a caption necessitates the utilization of both natural language processing and computer vision methodologies [49]. The utilization of the vision encoder-decoder model is conducive to the initialization of an image-to-text model, wherein a diverse array of pre-trained transformer-based vision models, such as [50]–[53], can be employed as the encoder. Similarly, the decoder component can be instantiated using a variety of pre-trained language models, including [46], [47], [54], [55]. For our study, we utilized the generative image-to-text transformer [56] that employed a CLIP/ViT-L/14 [57] for image encoding and a transformer network for decoding. This framework (**Figure 6**) was introduced as a method for consolidating multiple vision-language tasks, namely, image and video captioning, as well as question answering. It presents a straightforward model that relies on a single vision encoder and a single language decoder.

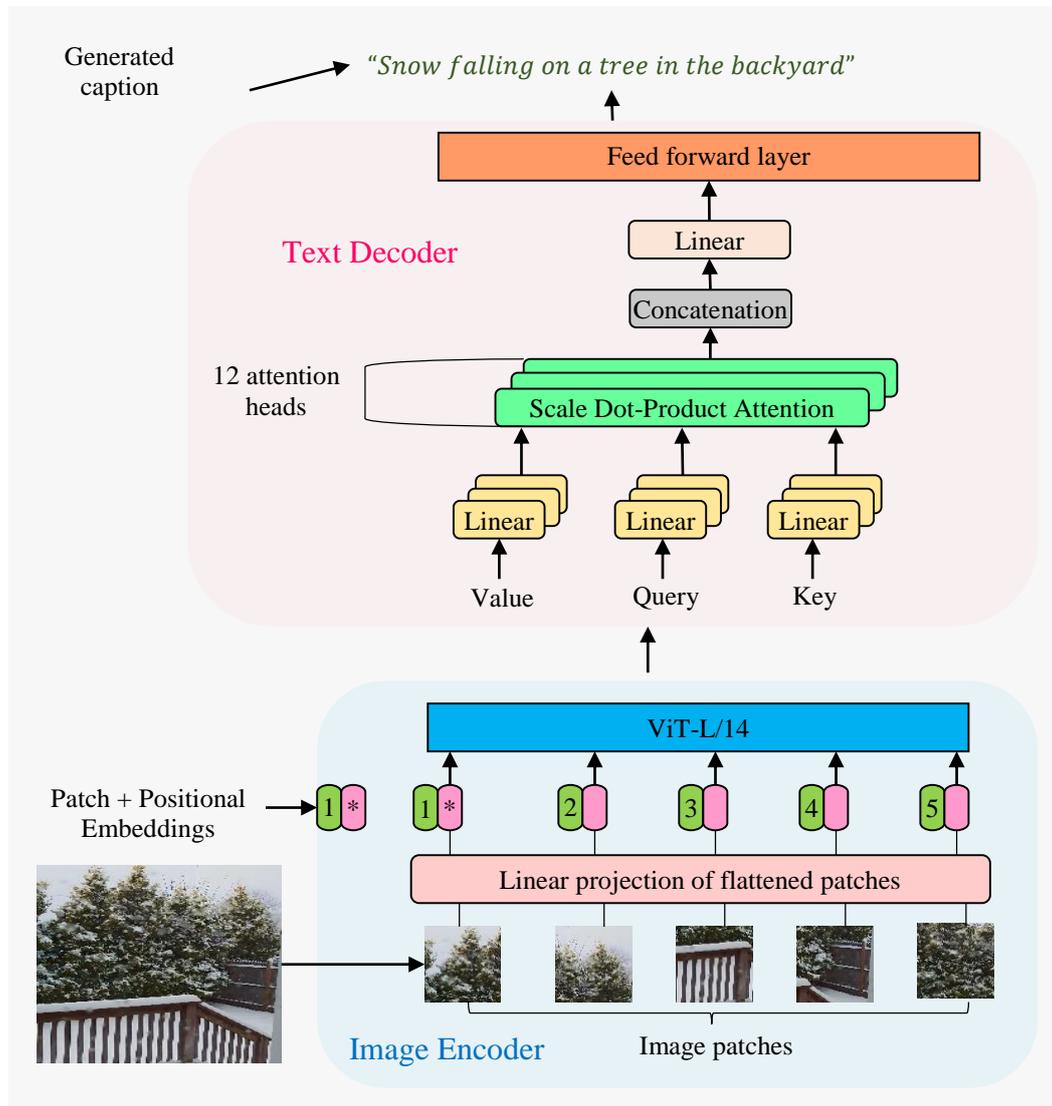

**Figure 6** Image captioning encoder-decoder architecture

The vision transformer is a computational framework designed for the purpose of image classification. The model employs an architecture inspired by transformers, which processes image patches. The process involves dividing an image into patches of a predetermined size. These patches are subsequently subjected to linear embedding, followed by the addition of position embeddings. The resultant vector sequence is then inputted into a conventional transformer encoder. $GIT_L$ employs the CLIP/ViT-L/14 vision and language model as its encoder that maps text and images to a shared vector space. CLIP models were designed to investigate the zero-shot generalization of models for image categorization tasks. The GPT-2 model is a transformer architecture that relies on an in-built masking paradigm to make certain that the predictions for a given token are based only on inputs received up to that point in the sequence. The model acquires an internal representation of the English language and is able to eliminate information from subsequent tokens in the prediction phase, allowing for the extraction of useful features for specific downstream applications. The model excels at its pretraining task, in particular, at producing textual outputs in response to inputs. Given its substantial size and the training on a dataset that is both extensive and varied, GPT-2 exhibits robust performance across a wide range of fields and data sets. In a manner comparable to GPT-2, CLIP has the capability to be employed in various visual classification benchmarks through the straightforward provision of the names of the visual categories that are to be identified, specifically in a zero-shot context. The image encoder, in this case, is a 347M parameter model pre-trained on 20M image-text pairs or 14 M images, which is a combination of COCO [58], SBU [59], CC3M [60], VG [61], and CC12M [62] datasets, for 30 epochs.

The text decoder is a 6-layer transformer network constituted by multiple transformer blocks, each with its own self-attention and feed-forward layers. The decoder network having 12 attention heads, and hidden dimension 768, is randomly initialized and operates by generating text descriptions in an auto-regressive manner. In the initial pretraining phase, the text undergoes tokenization and embedding processes. Subsequently, positional encoding is applied, followed by the inclusion of a norm layer. The transformer module is fed an input consisting of the concatenation of the image features and text embeddings. Beginning with a [BOS] token, the text is decoded until either [EOS] token is found or the maximum number of steps is reached. To facilitate reasoning across image features and text embeddings, a sequential-to-sequential attention mask is utilized. This attention mask ensures that each text token is influenced by the preceding tokens as well as all image tokens while the image tokens themselves attend to one another.

The text that has been extracted using captions contains a substantial amount of information from the image. This information can be utilized to generate supplementary discriminative features for the purpose of detecting sarcasm. In line with the initial textual input, the acquired additional information in the form of image captions also captures valuable knowledge for the purpose of sarcasm detection.

### 3.4.2 Image Caption Feature Representation

The caption feature representations are derived using the cross-lingual language model, just like the text feature representations as discussed in **section 3.2**. For the input sequence of caption words, $W' = \{w'_1, w'_2, w'_3, \ldots, w'_u\}$, the tokenized sequence is represented as depicted in ***Equation* (7)**.

$$W'_{tok} = \mathbb{X}_{tokenizer}(W') = \{[CLS], w'_1, w'_2, w'_3, \ldots, w'_u\} \tag{7}$$

where $w'_i \in \mathbb{R}^d$ and d denotes the embedding size. The sequential output, $C$, obtained for all tokens in the input $W'_{tok}$ can be denoted as shown in ***Equation* (8)**.

$$C = \mathbb{X}_{encoder}(W'_{tok}) \tag{8}$$

where, $C \in \mathbb{R}^{U \times d}$, where $d = 768$ stands for the dimension of each token, and U symbolizes the maximum total length of $W'_{tok}$.

### 3.5 Multi-level Cross-Domain Incongruity Representation

This section presents a thorough account of the methodologies proposed for capturing multi-level cross-domain discrepancies. The initial subsection focuses on the extraction of incongruity between feature representations obtained from textual and visual feature extraction branches. The subsequent subsection elucidates the modelling of discrepancies arising from the inclusion of the image caption modality.

### 3.5.1 Text-Image Incongruity

The relationship between each token pair of a sequence is taken into account in the internal representation generated using self-attention mechanism [63]. Given the inherent significance of discordance in sarcasm, the input tokens will place greater emphasis on image regions that directly contradict them. We take inspiration from this innate capacity for focused attention to create a layer, as illustrated in **Figure 7**, that can detect discrepancies between text and images. This layer in our model is designed to accept text features $S \in \mathbb{R}^{T \times d}$, as queries, and image features $I' \in \mathbb{R}^{r \times d}$, as keys and values. The model can then be directed to compare the query matrices of the text features to the key matrices of all image regions and allocate greater focus towards the ones that exhibit incongruity. The application of self-attention multiple times in parallel allows attending to different token pairs differently. Let us denote the total number of heads in the multi-head attention function as h. Then, for the i[th] attention head, the output $m_i$, can be computed as shown in **Equation (9)**.

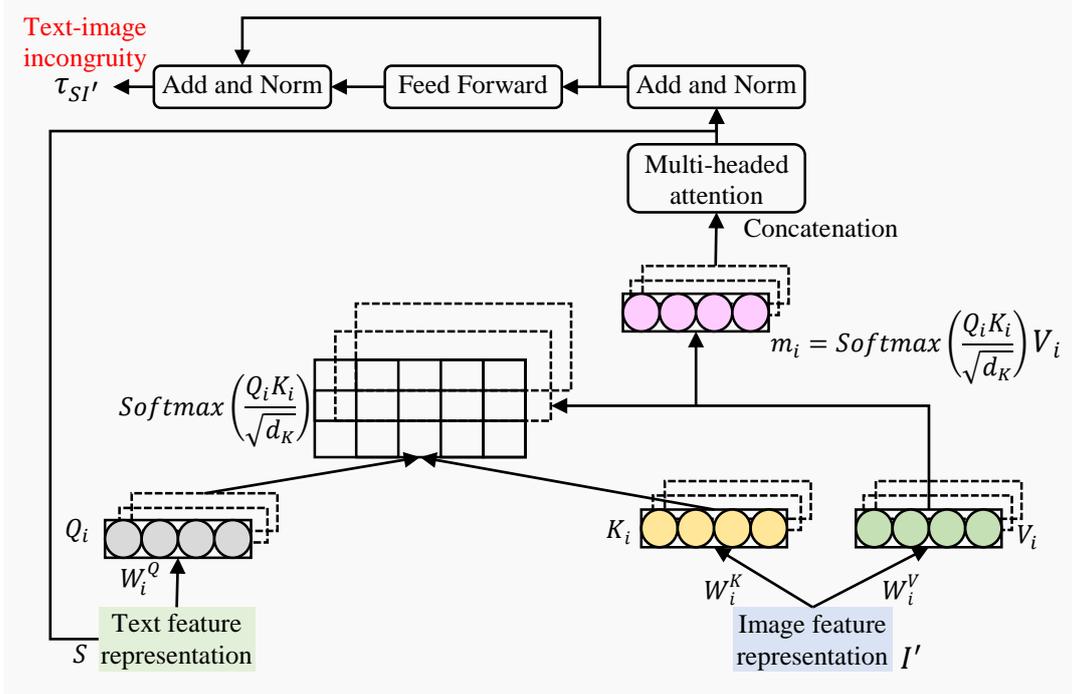

**Figure 7** Text-image incongruity

$$m_i(S, I) = Softmax(\frac{(SW_i^q)(I'W_i^k)^{\mathbb{T}}}{\sqrt{d_k}})(I'W_i^v) \tag{9}$$

where $d_k \in \mathbb{R}^{\frac{d}{h}}$, $m_i(S, I') \in \mathbb{R}^{T \times d_k}$, $\mathbb{T}$ symbolizes the matrix transpose operation, and $\{W_i^q, W_i^k, W_i^v\} \in \mathbb{R}^{d \times d_k}$ stand for the learnable query, key and value weight matrices. The multi-head attention is calculated using the output features of each head, $m_i$, as shown in **Equation (10)**.

$$Multi-head\ attention\ (S, I') = (m_1 \oplus m_2 \oplus m_3 \oplus \ldots \oplus m_h)W_O \tag{10}$$

where $W_O \in \mathbb{R}^{d \times d}$, and $\oplus$ denotes vector concatenation. Followed by a two-layer MLP and a residual connection, the text-image discrepancy captured can be obtained as illustrated in **Equation (11)**.

$$Incongurity\ (S, I') = LNorm\left(S + MLP\big(Multi-head\ attention\ (S, I')\big)\right) \tag{11}$$

where $Incongurity\ (S, I') \in \mathbb{R}^{T \times d}$. The final disparity representation, $\tau_{SI'} \in \mathbb{R}^d$ is obtained via **Equation (12)**.

$$\tau_{SI'} = [CLS] - encoding\ (Incongurity\ (S, I'))  \tag{12}$$

### 3.5.2 Text-Caption Incongruity

We utilize a co-attention mechanism, as demonstrated in the field of visual question answering by Lu et al. in [64], to effectively address the disparities between the original text and the generated image captions. As shown in **Figure 8**, the co-attention dual input comprises the feature representations of the text and the image captions, as obtained in ***Equation*** (**3**) and ***Equation*** (**8**), respectively. The initial step involves the computation of affinity matrix, $A \in \mathbb{R}^{T \times U}$ through the utilization of a bi-linear transformation $W$, as shown in ***Equation*** (**13**). The motive of this transformation is to accurately encapsulate the interrelation between the text and the captions of the accompanying images.

$$A = \tanh(SWC^{\mathbb{T}}) \tag{13}$$

Here, $S \in \mathbb{R}^{T \times d}$ denotes the text features, $C \in \mathbb{R}^{U \times d}$ denotes the image caption features (T and U denote the maximum size of these features correspondingly, and d symbolizes the hidden size of [44]), and $W \in \mathbb{R}^{d \times d}$ stands for a learnable parameter comprising weights. The affinity matrix that has been acquired serves to transform the attention space of the text into the attention space of the caption. In order to highlight the words that most significantly contribute to the incongruity characterizing the sarcastic expression, the affinity matrix is maximized over the text feature's locations to obtain caption attention, $a_c$, as illustrated in ***Equation*** (**14**).

$$a_c[u] = max_i(A_{i,u}) \tag{14}$$

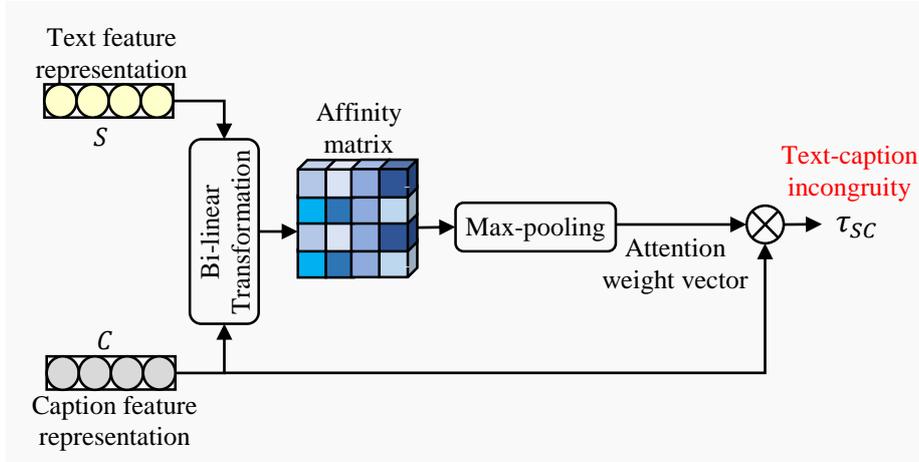

**Figure 8** Text-caption incongruity

The attention weight vector $v \in \mathbb{R}^U$ is thus calculated by applying a column-wised max-pooling operation kernel of dimensions $T \times 1$ over affinity matrix A, as depicted in ***Equation*** (**15**).

$$v = MaxPool\ (A) \tag{15}$$

The final text-image caption incongruity representation used to compute the model's prediction is obtained as $\tau_{SC} \in \mathbb{R}^d$, using ***Equation*** (**16**).

$$\tau_{SC} = vC \tag{16}$$

### 3.5.3 Final Fused Incongruity Representation and Model Prediction

After capturing both, the incongruity between the text and low-level image feature representation, $\tau_{SI'}$, and that between the text and high-level image description in the form of generated captions, $\tau_{SC}$, we concatenate the two disparity descriptors to obtain fusion vector, F for final model predictions, as shown in **Equation (17)**, where $\oplus$ signifies vector concatenation.

$$F = \tau_{SI'} \oplus \tau_{SC} \quad (17)$$

The fusion vector, F is put through the softmax function for classification purpose. The final prediction $y$ is computed as shown in **Equation (18)**, where $W_t \in \mathbb{R}^{2d}$ is a trainable parameter and b denotes the bias.

$$y = Softmax(W_t F + b) \quad (18)$$

## 4 Dataset Description

Our model for multimodal sarcasm detection is evaluated using two publicly available datasets constructed by Cai et al. [6] and Maity et al. [16].

### 4.1 Twitter Dataset (Cai et al. [6])

Every individual instance within this dataset comprises a combination of textual and corresponding visual information that has been acquired from the social media platform Twitter. Specific hashtag queries, such as #sarcasm, #sarcastic, #irony, #ironic, and so forth, were employed to gather positive samples which pertain to instances of sarcasm. Conversely, the negative samples, denoting non-sarcastic content, were obtained from tweets lacking the aforementioned hashtags. The dataset has been partitioned into three distinct subsets: the training set, which comprises 80% of the data; the development set, which accounts for 10%; and the test set, which also represents 10% of the dataset. The pertinent information has been presented in a tabular format, as outlined in **Table 2**.

**Table 2** Twitter multimodal sarcasm detection dataset statistics

|  | Training | Development | Testing |
| --- | --- | --- | --- |
| **Sarcastic** | 8,642 | 959 | 959 |
| **Non-Sarcastic** | 11,174 | 1,451 | 1,450 |
| **Total** | 19,816 | 2,410 | 2,409 |

### 4.2 MultiBully (Maity et al. [16])

The authors released a dataset collected from publicly available sources such as Twitter and Reddit to examine the influence of sentiment, emotion, and sarcasm in the detection of cyberbullying from multimodal memes within a code-mixed language context. They incorporated three supplementary tasks, namely sarcasm detection, sentiment analysis, and recognition of emotions, aiming to achieve better accuracy with respect to the primary objective, i.e. cyberbullying detection. For this study, we focus only on the subtask of sarcasm detection from multimodal data, i.e. text-image pairs. **Table 3** provides an overview of the statistics regarding the number of sarcastic and non-sarcastic memes held by the train, development, and test sets.

**Table 3** MultiBully dataset statistics

|  | Training | Development | Testing |
| --- | --- | --- | --- |
| **Sarcastic** | 1,545 | 201 | 429 |
| **Non-Sarcastic** | 2,552 | 384 | 743 |

| Total | 4,097 | 585 | 1,172 |

## 5 Experimental Results

This section provides elaborate information regarding the experimental settings and hyperparameters utilized in the proposed framework, as well as performance evaluations in comparison to baseline models.

### 5.1 Experimental Settings and Hyperparameters

Our model is implemented in PyTorch [65] using Google COLAB Pro Plus, which featured an NVIDIA V100 GPU, CUDA version 11.2, driver version 460.32.03, 40 GB of graphics memory, 100 GB of hard disk space, and 80 GB of RAM. We used softmax as a classifier for recognizing sarcasm in image-text pairs. The loss function employed to optimize the presented model was the L2-regularized binary cross-entropy, as represented in **_Equation_ (19)**.

$$J_{BCE}^{L2} = -\frac{1}{Train\ size}\sum_{i=1}^{Train\ size} y^i \log(\hat{y}_i) + (1-y^i)\log(1-\hat{y}_i) + \lambda \sum_{i=1}^{Train\ size} w_i^2 \quad (19)$$

In the present study, the Adam optimizer [66] was employed as the optimization algorithm to update all the trainable parameters of the model. The constituted model was trained with a learning rate of $5 \times 10^{-5}$, using a batch size of 32. Additionally, a warmup rate of 0.1 was used to control the variation of learning rate and facilitate the optimization process. To mitigate the issue of overfitting, we implemented the dropout technique with a probability value of 0.5. Further, to save time and computational resources, an early stopping patience value of 5 was enforced. By employing this technique, we were able to effectively monitor the training process and terminate it once the model's performance began to plateau, thereby ensuring that the model does not become overly specialized to the training data and could exhibit good generalization capabilities towards novel data. The proposed model was fine-tuned for 15 epochs.

### 5.2 Evaluation Metrics

In accordance with the evaluation methodology employed by [6], we employ four widely used performance metrics, namely Accuracy (A), Precision (P), Recall (R), and F1-score (F1), to evaluate the efficacy of the proposed model. The mathematical formulae of these metrics are provided in **Table 4**.

**Table 4** Performance metrics

| Performance metric | Mathematical formula |
|---|---|
| Accuracy ($A$) | $\frac{True\ Pos. + True\ Neg.}{True\ Pos. + True\ Neg. + False\ Pos. + False\ Neg.}$ |
| Precision ($P$) | $\frac{True\ Pos.}{True\ Pos. + False\ Pos.}$ |
| Recall ($R$) | $\frac{True\ Pos.}{True\ Pos. + False\ Neg.}$ |
| F-1 Score ($F1$) | $\frac{2 \times P \times R}{P + R}$ |

### 5.3 Baselines
The presented framework is being compared to the baselines that are discussed below.
**Text-only**

These models made the final predictions solely on the basis of textual data, including **TextCNN** [72], **SIARN** [68], **SMSD** [69], **MIARN** [68], **BiLSTM** [70], **BERT** [47], and **TextGraph** [37].

**Visual-only**

These models utilized only the visual information for detecting sarcasm, including **Image** (vectors obtained after ResNet pooling layer) [6], **ViT** [51], **ResNet-152** [71], and **ImageGraph** [37].

**Multimodal**

The models mentioned in the text incorporate both text and image data, and they have the capability to incorporate additional inputs from other modalities, depending on the specific framework being used. The **HFM** [6] framework was designed to combine text, image, and attribute features in a hierarchical manner. The **Bridge** [10] model utilized a bridge layer to map the image feature representation into the BERT vector space. It also addressed the differences between images and text by incorporating multi-head attention. The **Res-BERT** [67] concatenated both text and image features to detect sarcasm, while **Att-BERT** [67] employed separate attention modules to capture disparities within and between different modalities. The **D&R Net** [8] approach involved incorporating ANPs, text, and images to analyze the contextual contrast between text and images, as well as the level of semantic association. The Coupled Attention framework, introduced in **CAN** [11], aims to prioritize certain keywords and regions in text and image data. The authors of **InCrossMGs** [37] utilized graph convolution networks to analyze the sentiment incongruity present in text and images. They achieved this by constructing both in-modal and cross-modal graphs. In their study, **CMGCN** [38] employed object recognition techniques to identify important sections of an image. They then established connections between textual tokens and these identified parts to effectively detect sarcasm. **MuLOT** [12] captures internal relationships within a single modality by employing self-attention. Additionally, it leverages optimal transport techniques to take advantage of cross-modal associations. Logic-based techniques were utilized in order to identify occurrences of multimodal rumors and sarcasm in **LogicDM** [39].

## 5.4 Experimental Results

**Table 5 and Table 6** present a comparative analysis between our proposed methodology and other cutting-edge benchmarks. Our proposed solution presented in this study demonstrates superior performance compared to the existing state-of-the-art approaches across each of the four evaluation metrics, namely $Accuracy$, $Recall$, $Precision$, and $F1-score$. Our model indicates a significant enhancement of 2.07% in $Accuracy$ and 0.51% in $F1-score$ compared to the current state-of-the-art MuLOT (with OCR) [12] for Twitter dataset and 1.49% in $Accuracy$ and 0.47% in $F1-score$ compared to Maity et al. [16] for MultiBully dataset. This outcome serves as empirical evidence supporting the efficacy of our proposed model.

It can be observed from **Table 5** and **Table 6** that the independent treatment of images or text does not yield satisfactory results in addressing the sarcasm detection problem. From a qualitative perspective, it can be observed that models relying solely on images tend to exhibit lower performance compared to models relying solely on text. This is due to the fact that an image alone lacks the necessary information to accurately identify the presence of sarcasm. Furthermore, it is evident that unimodal approaches exhibit inferior performance in sarcasm identification, highlighting the need for multimodal approaches to be employed for more effective sarcasm detection. It can be inferred from **Table 5** and **Table 6** that our proposed approach exhibits superior performance in comparison to other multimodal approaches by effectively capturing the disparities between the image and text pairs. **Figure 9** and **Figure 10** depict a visual representation for the comparative analysis between our proposed framework and other multimodal methods based on various evaluation metrics.

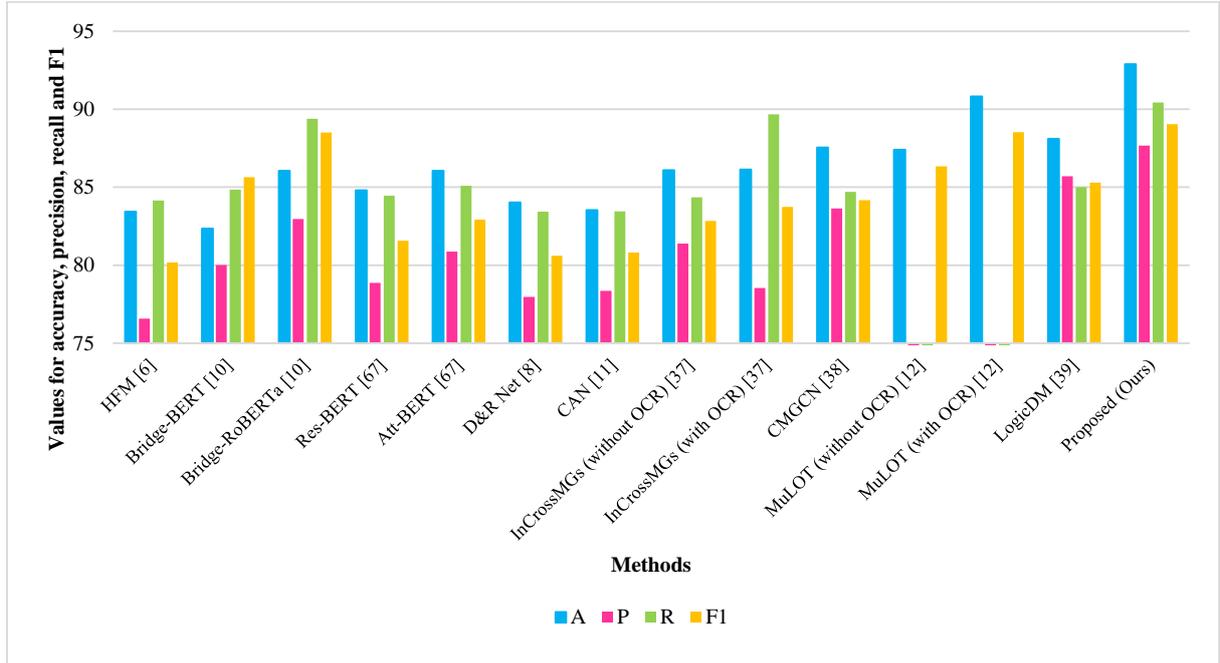

**Figure 9** Comparison of accuracy, precision, recall and F1 of our proposed model with the baseline approaches for multimodal Twitter dataset

**Table 5** Performance comparison with state-of-the-art unimodal and multimodal baselines for Twitter multimodal sarcasm detection dataset

| Modality | Method | A | P | R | F1 |
|---|---|---|---|---|---|
| Text | TextCNN [72] | 80.03 | 74.29 | 76.39 | 75.32 |
| | SIARN [68] | 80.57 | 75.55 | 75.70 | 75.63 |
| | SMSD [69] | 80.90 | 76.46 | 75.18 | 75.82 |
| | MIARN [68] | 82.48 | 79.67 | 75.18 | 77.36 |
| | BiLSTM [70] | 81.90 | 76.66 | 78.42 | 77.53 |
| | BERT [47] | 83.85 | 78.27 | 82.27 | 80.22 |
| | TextGraph [37] | 84.23 | 78.86 | 82.48 | 80.63 |
| | **Cross-lingual language model (Ours)** | **88.32** | **87.03** | **87.61** | **87.32** |
| Image | Image [6] | 64.76 | 54.41 | 70.80 | 61.53 |
| | ViT [51] | 67.83 | 57.93 | 70.07 | 63.43 |
| | ResNet-152 [71] | 72.60 | 65.11 | 67.15 | 66.11 |
| | ImageGraph [37] | 73.89 | 63.24 | **82.17** | **71.47** |
| | **Self-regulated ConvNet + Lightweight Attention (Ours)** | **74.11** | 63.62 | 72.98 | 67.98 |
| Multimodal | HFM [6] | 83.44 | 76.57 | 84.15 | 80.18 |
| | Bridge-BERT [10] | 82.35 | 80.01 | 84.84 | 85.64 |
| | Bridge-RoBERTa [10] | 86.05 | 82.95 | 89.39 | 88.51 |
| | Res-BERT [67] | 84.80 | 78.87 | 84.46 | 81.57 |
| | Att-BERT [67] | 86.05 | 80.87 | 85.08 | 82.92 |
| | D&R Net [8] | 84.02 | 77.97 | 83.42 | 80.60 |
| | CAN [11] | 83.53 | 78.36 | 83.45 | 80.82 |
| | InCrossMGs (without OCR) [37] | 86.10 | 81.38 | 84.36 | 82.84 |
| | InCrossMGs (with OCR) [37] | 86.13 | 78.54 | 89.68 | 83.74 |
| | CMGCN [38] | 87.55 | 83.63 | 84.69 | 84.16 |
| | MuLOT (without OCR) [12] | 87.41 | - | - | 86.33 |
| | MuLOT (with OCR) [12] | 90.82 | - | - | 88.52 |
| | LogicDM [39] | 88.10 | 85.70 | 85.00 | 85.30 |
| | **Proposed (Ours)** | **92.89** | **87.67** | **90.43** | **89.03** |

**Table 6** Performance comparison with state-of-the-art unimodal and multimodal baselines for MultiBully dataset

| Modality | Method | *A* | *P* | *R* | *F1* |
|---|---|---|---|---|---|
| Text | m-BERT-GRU [16] | 59.72 | - | - | 59.12 |
| | **Cross-lingual language model (Ours)** | **63.83** | **61.72** | **62.69** | **62.20** |
| Image | ResNet [16] | 59.39 | - | - | 57.79 |
| | **Self-regulated ConvNet + Lightweight Attention (Ours)** | **62.91** | **61.07** | **61.98** | **61.53** |
| Multimodal | Maity et al. [16] | 62.99 | - | - | 63.80 |
| | **Proposed (Ours)** | **64.48** | **63.69** | **64.87** | **64.27** |

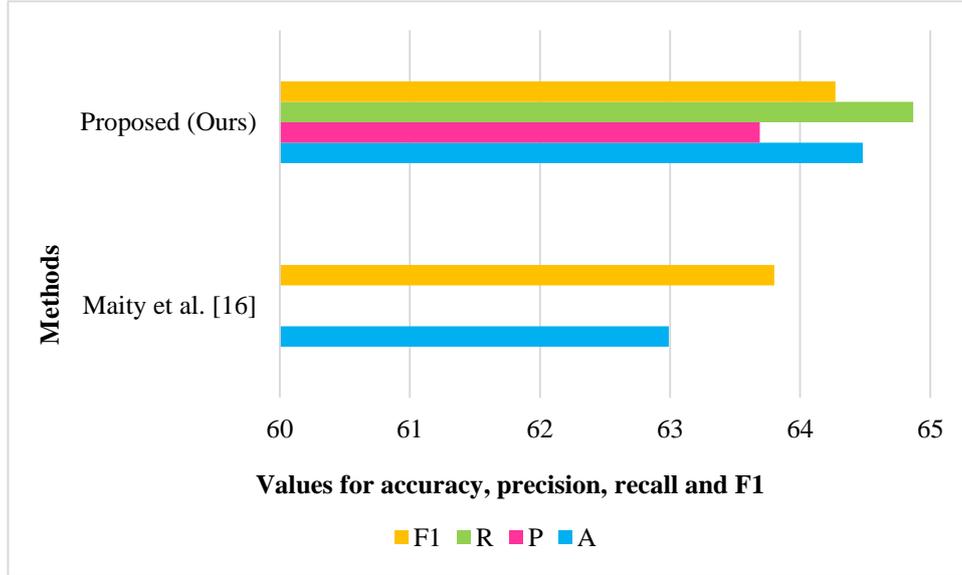

**Figure 10** Comparison of accuracy, precision, recall and F1 of our proposed model with the baseline approaches for MultiBully dataset

## 5.5 Ablation Study

In this section, a set of ablation trials are carried out on the two datasets, namely the Twitter multimodal sarcasm dataset and the MultiBully dataset. The objective is to thoroughly evaluate the effectiveness of each proposed component for the multimodal input. As per the suggested framework, we have omitted the lightweight spatially aware attention module from the attentional visual feature extraction branch, referred to as the *proposed model w/o visual attention*. Additionally, we have also eliminated the incongruity representation $\tau_{SI'}$ from the proposed model, denoted as the *proposed model w/o $\tau_{SI'}$*. Likewise, the removal of $\tau_{SC}$ from the proposed model has been represented as *proposed model w/o $\tau_{SC}$*. The results obtained from the ablation trials are succinctly displayed in **Table 7**.

**Table 7** Ablation Study

| Approaches | Twitter multimodal sarcasm detection dataset | | | | MultiBully dataset | | | |
|---|---|---|---|---|---|---|---|---|
| | *A* | *P* | *R* | *F1* | *A* | *P* | *R* | *F1* |
| *Proposed model w/o visual attention* | 89.33 | 85.16 | 87.42 | 86.28 | 62.02 | 59.14 | 59.86 | 59.49 |
| *Proposed model w/o $\tau_{SI'}$* | 80.12 | 79.68 | 80.83 | 80.25 | 52.04 | 50.86 | 51.94 | 51.39 |
| *Proposed model w/o $\tau_{SC}$* | 83.67 | 82.04 | 82.99 | 82.51 | 53.89 | 49.10 | 51.6 | 50.32 |
| **Proposed (Ours)** | 92.89 | 87.67 | 90.43 | 89.03 | 64.48 | 63.69 | 64.87 | 64.27 |

From **Table 7** it can be observed that the exclusion of the attention module from the visual branch of our proposed model hinders the overall predictive capability. Also, the elimination of incongruity representations also leads to a decline in the model's performance, indicating that the inclusion of multi-head and co-attention to capture incongruity between textual and image features is crucial for the detection of sarcasm. Based on the results mentioned above, it can be inferred that our proposed model demonstrates superior ability in capturing the disparity between visual and textual modalities, consequently facilitating effective acquisition of the underlying sarcasm.

# 6   Conclusion and Future Scope

In this study, we propose a novel approach for multimodal sarcasm recognition by utilizing image captions to create a multi-tier cross-modal incongruity representation. The inclusion of captions in this particular context provides a valuable opportunity to improve our understanding of the meaning conveyed by images and draw attention to the difference in emotional tone between the message conveyed in a tweet and the visual content that accompanies it. This methodology improves the efficacy of the suggested framework in identifying instances of sarcasm within two openly available datasets. The experimental results indicate that our proposed framework consistently improves upon state-of-the-art baselines for multimodal sarcasm detection. The effectiveness of each module within the provided framework is additionally supported by the ablation study. In future investigations, our objective is to analyze the feasibility of adapting our model to accommodate supplementary multimodal tasks. Future research endeavors may delve into improved fusion methodologies in order to better capture the interconnections among various modalities. Furthermore, researchers may also aim to integrate low-parameter or domain-adaptive models into their studies. This is due to the fact that models with a reduced number of trainable parameters provide notable benefits by facilitating real-time deployment.